\documentclass[10pt,twocolumn,letterpaper]{article}

\usepackage{iccv}
\usepackage{times}
\usepackage{epsfig}
\usepackage{graphicx}
\usepackage{amsmath}
\usepackage{amssymb}

\usepackage{graphicx}
\usepackage{amsmath}
\usepackage{amssymb}
\usepackage{booktabs}
\usepackage{multirow}
\usepackage{pifont}
\usepackage{amsmath,scalerel}
\DeclareMathOperator*{\concat}{\scalerel*{\Vert}{\sum}}

\usepackage{pifont}
\usepackage{color}
\usepackage[table]{xcolor}
\usepackage{colortbl}

\definecolor{yellow}{rgb}{1,1, 0.6}
\definecolor{orange}{rgb}{1, 0.8, 0.6}

\usepackage[pagebackref=true,breaklinks=true,letterpaper=true,colorlinks,bookmarks=false]{hyperref}

\usepackage[capitalize]{cleveref}
\crefname{section}{Sec.}{Secs.}
\Crefname{section}{Section}{Sections}
\Crefname{table}{Table}{Tables}
\crefname{table}{Tab.}{Tabs.}
 \iccvfinalcopy 

\def\iccvPaperID{2697} 

\ificcvfinal\fi

\begin{document}

\title{Leveraging Spatio-Temporal Dependency for Skeleton-Based Action Recognition}

\author{Jungho Lee$^{1}$
,
Minhyeok Lee$^{1}$
,
Suhwan Cho$^{1}$
,
Sungmin Woo$^{1}$
,
Sungjun Jang$^{1}$
,
Sangyoun Lee$^{1}$\\
\vspace{0.01cm}\\
$^{1}$Yonsei University\\
{\tt\small \{2015142131, hydragon516, chosuhwan, smw3250, jeu2250, syleee\}@yonsei.ac.kr}
}


\maketitle
\ificcvfinal\thispagestyle{empty}\fi

\begin{abstract}
	Skeleton-based action recognition has attracted considerable attention due to its compact representation of the human body's skeletal sructure. Many recent methods have achieved remarkable performance using graph convolutional networks (GCNs) and convolutional neural networks (CNNs), which extract spatial and temporal features, respectively. Although spatial and temporal dependencies in the human skeleton have been explored separately, spatio-temporal dependency is rarely considered. In this paper, we propose the Spatio-Temporal Curve Network (STC-Net) to effectively leverage the spatio-temporal dependency of the human skeleton. Our proposed network consists of two novel elements: 1) The Spatio-Temporal Curve (STC) module; and 2) Dilated Kernels for Graph Convolution (DK-GC). The STC module dynamically adjusts the receptive field by identifying meaningful node connections between every adjacent frame and generating spatio-temporal curves based on the identified node connections, providing an adaptive spatio-temporal coverage. In addition, we propose DK-GC to consider long-range dependencies, which results in a large receptive field without any additional parameters by applying an extended kernel to the given adjacency matrices of the graph. Our STC-Net combines these two modules and achieves state-of-the-art performance on four skeleton-based action recognition benchmarks. Code is available at \href{https://github.com/Jho-Yonsei/STC-Net/}{https://github.com/Jho-Yonsei/STC-Net}.
\end{abstract}

\section{Introduction}
\label{sec:intro}
Action recognition is one of the most important video understanding tasks used in various applications such as virtual reality and human--computer interaction. Recent studies on action recognition are divided into two methods, RGB-based~\cite{wang2016temporal,veeriah2015differential} and skeleton-based methods~\cite{yan2018spatial,shi2019skeleton,shi2019two,chen2021channel,cheng2020decoupling,lee2022hierarchically}. Action recognition using the skeleton modality receives a video sequence with the three-dimensional coordinates of major human joints as its input. Skeleton-based action recognition has the advantage of being able to create a lightweight model with low computational complexity by compactly compressing the structure of the human body. In addition, it has the benefit of robustness in that it is not affected by background noise, weather, and lighting conditions unlike RGB-based methods.

\begin{figure}[t]
	\centering
	\includegraphics[width=\linewidth]{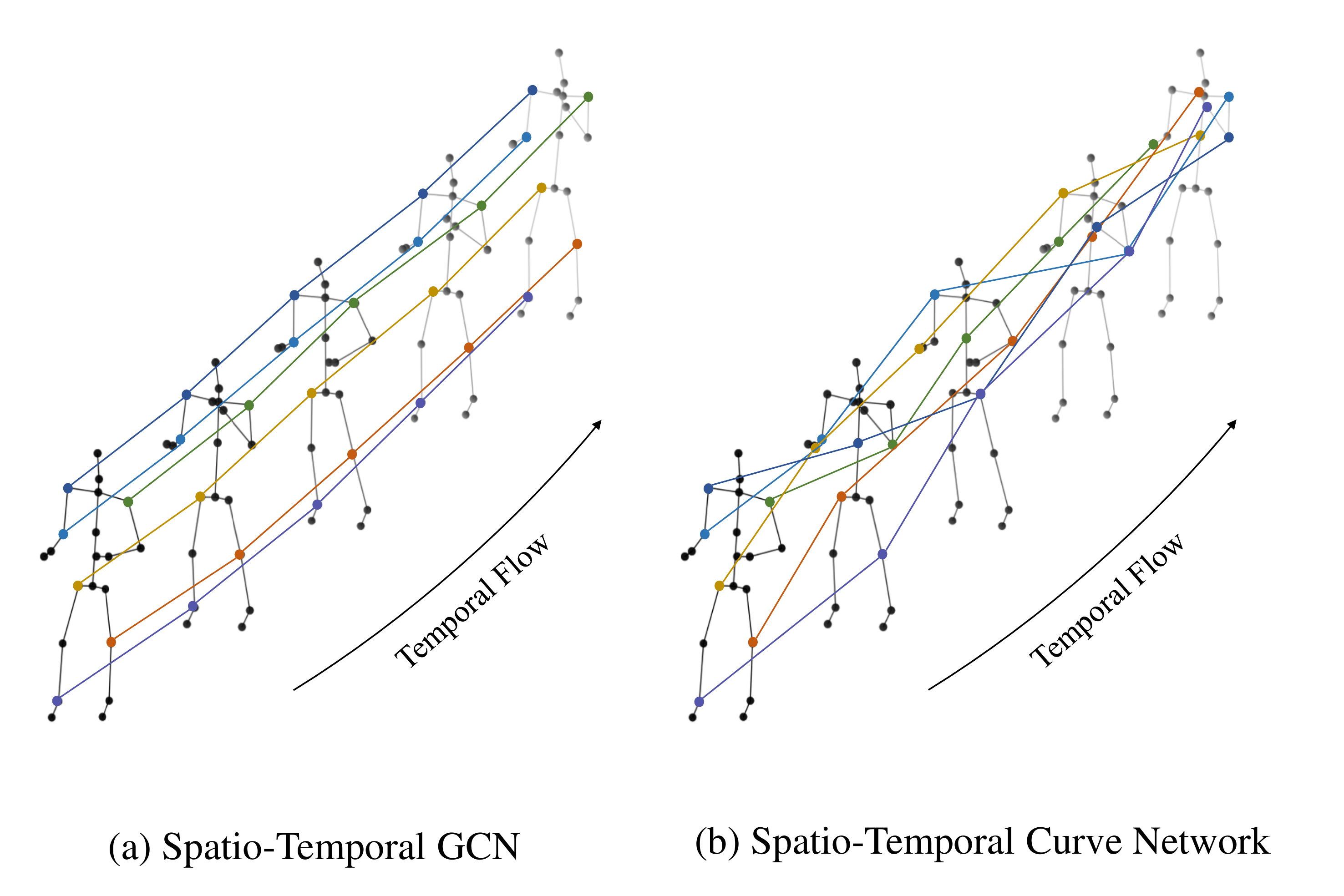}
	\caption{Comparison of temporal flows of spatio-temporal GCN (a) and STC module of our model (b). (b)'s curves have adaptive spatio-temporal receptive field by aggregating different nodes for different frames, whereas (a) treats the temporal features of each node independently.}
	\vspace{-5mm}
	\label{fig:curvenet}
\end{figure}

Earlier approaches~\cite{duvenaud2015convolutional,hamilton2017inductive,ying2018hierarchical,kipf2018neural,monti2017geometric} extract features by dealing with every joint independently, which means that they do not consider information between structurally correlated human joints. However, the connections between human joints are identified as a graph structure after Yan~\etal~\cite{yan2018spatial} has proposed spatio-temporal graph convolutional networks (GCNs) for the skeleton modality. Recent approaches~\cite{shi2019two,cheng2020decoupling,chen2021channel,chi2022infogcn} adopt the GCNs as their baseline and attempt to enlarge the receptive field on the spatial domain.

However, methods based on Yan~\etal's GCNs have several limitations. (1) When a person performs an action, the movement of their body parts occurs in both space and time, and these two aspects of the movement are inherently interconnected. While incorporating both spatial and temporal components can provide a more complete and accurate representation of human actions, it is not feasible to directly utilize this spatio-temporal interconnectivity as the spatial and temporal modules exist independently of each other. (2) As they use graphs that include only the connectivity of physically adjacent joints, their networks with such graphs have small spatial receptive fields. Although several self-attention methodologies~\cite{shi2019two,chen2021channel} have been proposed to increase the spatial receptive field, they still rely on using physically adjacent graphs, which can lead to biased results towards those physically adjacent graphs and highlight a potential limitation in their effectiveness. To handle this problem, Liu~\etal~\cite{liu2020disentangling} has proposed a multi-scale graph that identifies the relationship between structurally distant nodes. However, as stated by Yan~\etal~\cite{yan2018spatial}, although it is crucial to differentiate human motion into concentric and eccentric patterns, Liu~\etal's method does not account for such patterns. Additionally, Liu~\etal's model suffers from the limitation of having high model complexity, as there are too many operations parallelly existing in a single layer.

To solve limitation (1), we propose a Spatio-Temporal Curve (STC) module to reflect direct spatio-temporal dependencies in a skeleton sequence. In addition to applying temporal convolution to aggregate node-wise sequential features, we construct curves that consider the sequential spatio-temporal features for every node and aggregate them with input feature map. To create the curves, we choose the most highly correlated nodes in feature space between all adjacent frames and connect them. Therefore, a more semantically effective graph structure can be adaptively generated by giving autonomy to the temporal connections between nodes.~\cref{fig:curvenet} compares the temporal flows of the spatio-temporal GCN for existing methods~\cite{yan2018spatial,shi2019two,chen2021channel} and our proposed method.~\cref{fig:curvenet} (a) shows that the model reflects only the features of the same nodes in every frame, while~\cref{fig:curvenet} (b) shows that the model considers the spatio-temporal correlations through the generated curves that take account of features of different nodes in adjacent frames. Inspired by~\cite{Xiang_2021_ICCV}, we use an aggregation module to effectively combine all the curve features and apply them to the input feature map.

To handle limitation (2), we propose Dilated Kernels for Graph Convolution (DK-GC) to have large spatial receptive field for skeletal modality without any additional parameters. The GCNs for the human skeleton aggregate inward-facing (centripetal), identity, and outward-facing (centrifugal) features, unlike convolutional neural networks (CNNs), which aggregate left, identity, and right pixels features. To apply the dilated kernel to such GCNs, we create adjacency matrices to identify structurally distant relationships by modifying centripetal and centrifugal matrices. To incorporate spatial receptive fields from low-level to high-level, we divide the spatial module into several branch operations with different dilated windows. Meanwhile, dilated graph convolution has already been introduced by Li~\etal~\cite{li2019deepgcns} for 3D point clouds analysis task. However, Li~\etal's dilated graph convolution is completely different from what we propose and is not suitable for human skeletal modality. Firstly, this method does not utilize the given adjacency matrices, but instead uses dynamic graph via k-nearest neighborhood (k-NN) algorithm. The inability to utilize the given adjacency matrices reduces the robustness for the action recognition model as Shi~\etal~\cite{shi2019two} experimentally has proven that not using those matrices leads to inferior performance. Moreover, the k-NN alone cannot identify all the physically adjacent nodes. The second reason is that Li~\etal's method requires a lot of GPU resources. It causes very high GPU memory consumption and low inference speed since the dynamic graphs are constructed by computing all pairwise distances between all the nodes for every GCN layer. 

To verify the superiority of our STC-Net, extensive experiments are conducted on four skeleton-based action recognition benchmark datasets: NTU-RGB+D 60~\cite{shahroudy2016ntu}, NTU-RGB+D 120~\cite{liu2019ntu}, Kinetics-Skeleton~\cite{kay2017kinetics}, and Northwestern-UCLA~\cite{wang2014cross}.

Our main contributions are summarized as follows:

\begin{itemize}
	\item [$\bullet$] We propose the Spatio-Temporal Curve (STC) module to leverage the direct spatio-temporal correlation between different nodes of different frames.
	\item [$\bullet$] We propose the Dilated Kernels for Graph Convolution (DK-GC) that makes the model have a large spatial receptive field without any additional parameters by modifying the given skeletal adjacency matrices.
	\item [$\bullet$] Our proposed STC-Net outperforms existing state-of-the-arts methods on four benchmarks for skeleton-based action recognition.
\end{itemize}

\begin{figure*}[t]
	\centering
	\includegraphics[width=\linewidth]{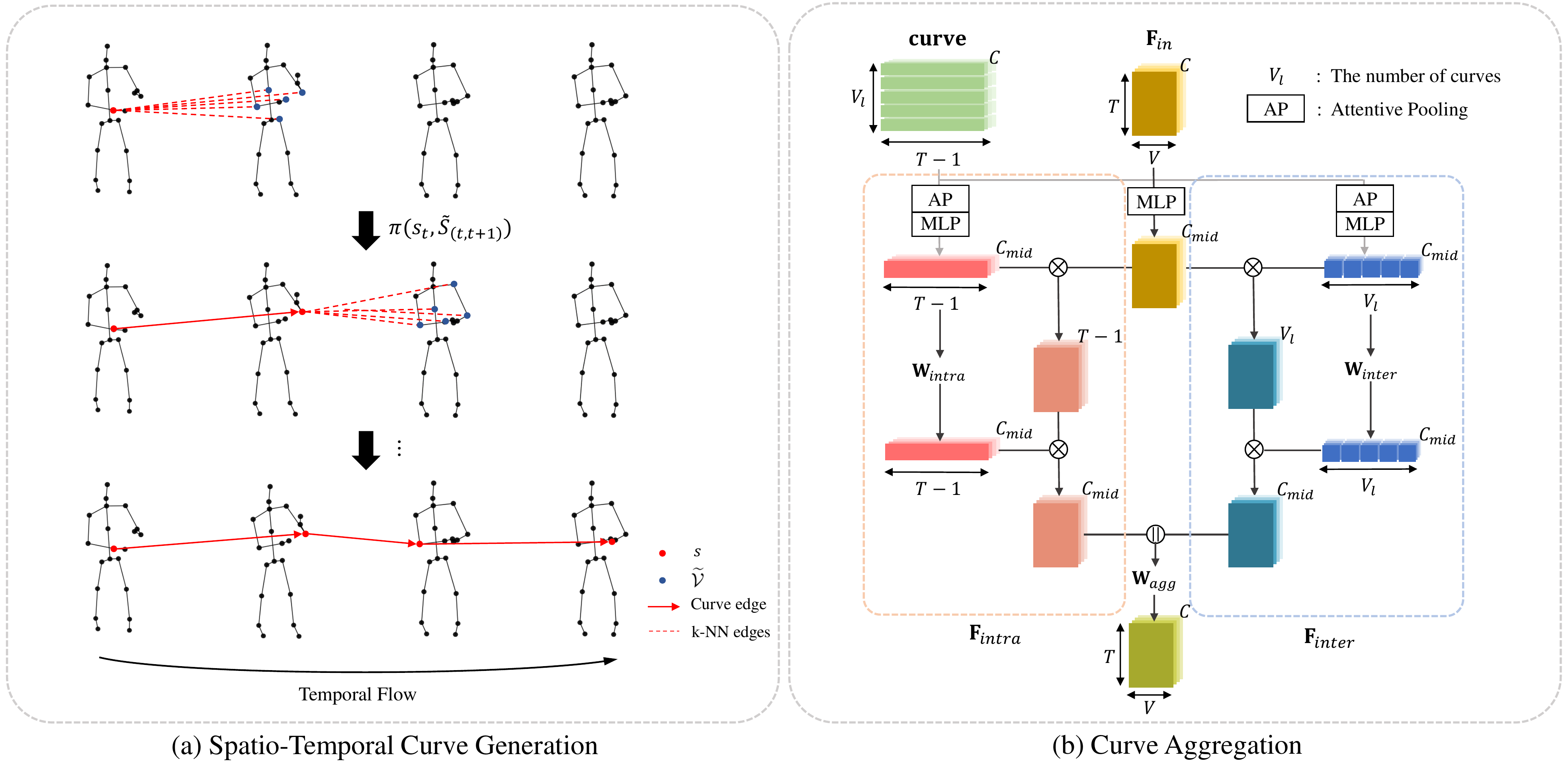}
	\vspace{-6mm}
	\caption{(a) Process of constructing an spatio-temporal curve and (b) the curve aggregation module that applies the curves to the input feature map $\mathbf{F}_{\textrm{in}}$. The dotted line in (a) is the relationship between the query node and the key nodes extracted by the inter-frame k-NN. The node selection policy $\pi$ adopts the node of the highest correlation score with the query node as the next point in the curve. $||$ and $\times$ denote concatenation and matrix multiplication, respectively.}
	\vspace{-2mm}
	\label{fig:curves}
\end{figure*}

\section{Related Work}
\label{sec:related}
\subsection{Skeleton-Based Action Recognition}
Previous skeleton-based action recognition methods~\cite{duvenaud2015convolutional,kipf2018neural,niepert2016learning} do not consider the relationships between joint nodes for the human skeleton but treat all the joint nodes independently. However, Yan~\etal~\cite{yan2018spatial} treats this modality as a graph structure. Most recent methods~\cite{shi2019two,liu2020disentangling,shi2020skeleton,chen2021channel} tend to rely on GCNs to deal with spatial correlation. They reflect the subordinate correlations of nodes by identifying the relationships between adjacent joint nodes. In particular, models with graph structures and a self-attention mechanism~\cite{shi2019skeleton,chen2021channel} show remarkable performance. In addition, many RNN-based and CNN-based temporal modules have been proposed to deal with action recognition sequence data defined as time series.

However, there are drawbacks to these methods, and the first is that structurally distant nodes are treated almost independently because of their small spatial receptive fields, because the given graph considers only the adjacent relationships of nodes. To treat this limitation, learnable graph structures with self-attention mechanisms~\cite{shi2019two,chen2021channel,chi2022infogcn} are proposed. However, these structures use not only the graphs with self-attention mechanisms, but also given graph with adjacent connections. In other words, even if the receptive fields of their models are enlarged with self-attention mechanisms, they do not highlight the appropriate nodes much because they are highly biased to the given graph. Second, although several methods for handling skeleton sequences~\cite{liu2017skeleton,liu2017global} have been proposed, most of them do not take account of spatio-temporal dependencies because they consider only node-wise sequences. In other words, most recent methods are vulnerable to this dependency because of their low spatio-temporal receptive fields.

\subsection{Curves for 3D Point Clouds}
3D point clouds are unstructured representations of 3D coordinates, tasks for them are to analyze the information for 3D points that exist without any adjacency matrices given. In order to effectively analyze these point clouds, Xiang~\etal~\cite{Xiang_2021_ICCV} propose CurveNet, which identifies the relationships between every 3D points by aggregating both local and non-local features. In one scene, CurveNet initializes $n$ starting points and constructs $n$ curves with length $l$ by finding the most correlated points. Applying the curves to the feature space, both the local and non-local point connections are identified. We propose an spatio-temporal curve module that makes the curves applicable to skeletal video data, and effectively increases the spatio-temporal receptive field.

\section{Methodology}
\label{sec:method}


\subsection{Spatio-Temporal Curve Module}
\label{sec:ifc}

In this subsection, we describe the spatio-temporal curve (STC) module to consider the direct spatio-temporal correlations of skeleton sequences.

\paragraph{\bf{Spatio-Temporal Curve Generation.}}
\label{subsec:ifc}

In order to construct a curve between frame $t$ and $(t+1)$, it is essential to find and connect the key node $\widetilde{v}_{t+1}$ that is most semantically close to the query node $v_{t}$, where $v_{t}$ is a specific node on frame $t$. To specify the node $\widetilde{v}_{t+1}$, we select $k$ nodes in frame $(t+1)$ that are semantically close to the query node on the feature space. The nodes are selected through inter-frame k-NN, which applies Euclidean distance-based k-NN algorithms between all adjacent frames to obtain the semantically closest $k$ nodes in the feature space. However, if the query node $v_{t}$ and obtained key node $\widetilde{v}_{t+1}$ refer to structurally identical locations, the model may identify only the same nodes between adjacent frames. Therefore, it hinders the model's ability to capture diverse curves. To prevent this problem, we apply the k-NN algorithm while excluding the node of frame $(t + 1)$ that is located in the same structural position as query node $v_{t}$. The proposed inter-frame k-NN is as follows:
\begin{equation}
	\widetilde{\mathcal{V}}_{(t, t+1)} = \sum_{v \in V} {\textrm{k-NN}(v_{t}, \mathcal{V}_{t+1} - \{v_{t+1}\})},
\end{equation}
where $\widetilde{\mathcal{V}}_{(t,t+1)}$ denotes a set of $k$ nodes in frame $(t+1)$ that are the semantically closest nodes to the query node $v_{t}$, and $(\mathcal{V}_{(t+1)}-\{v_{t+1}\})$ refers to the node set in frame $(t+1)$ except the node that is structurally the same as $v_{t}$.

To construct effective curves, the key node most highly correlated with the query node $v_{t}$ should be extracted using the node set $\widetilde{\mathcal{V}}_{(t,t+1)}$ created by inter-frame k-NN. For this extraction process, we propose an extended method for the node selection policy $\pi$ in~\cite{Xiang_2021_ICCV} to choose the key node $\widetilde{v}_{t+1}$. The policy in~\cite{Xiang_2021_ICCV} reflects only nodes in a single frame without considering the temporal feature space. To consider the time domain as well, we apply a new node selection policy $\pi_{t}$ to choose the key node by reflecting the key node features and semantically adjacent node set $\widetilde{\mathcal{V}}_{(t, t+1)}$.
\begin{equation}
	s_{t+1} = \pi_{t}\left(s_{t}, \widetilde{\mathcal{S}}_{(t, t+1)}\right),\  1 \leq t \in \mathbb{Z}^{+} \leq T-1,
\end{equation}
where $s_{t}$ and $s_{t+1}$ refer to the embedded spaces of the query and the key nodes, $\widetilde{\mathcal{S}}_{(t, t+1)}$ denotes the embedded features of $\widetilde{\mathcal{V}}_{(t,t+1)}$, and $T$ is the length of the skeleton sequence, which is equal to $(curve\ length + 1)$. We build a learnable policy $\pi_{t}$ to consider both the features of query node $v_{t}$ and the features of extracted node set $\widetilde{\mathcal{V}}_{(t, t+1)}$. We obtain a new agent feature map $\mathbf{M}_{\textrm{agent}}$ by passing those features through an agent MLP layer. Then, we extract the node with the highest correlation score in adjacent frame $(t+1)$ via $\mathbf{M}_{\textrm{agent}}$. We obtain $s_{t+1}$ from the input feature map $\mathbf{F}_{\textrm{in}}$ via $\widetilde{v}_{t+1}$, and it becomes the next waypoint of the curve. The process to choose the key node is as follows:

\begin{equation}
	\textrm{M}_{\textrm{agent}} = \textrm{MLP}_{\textrm{agent}}\left(s_{t}\parallel\widetilde{\mathcal{S}}_{(t, t+1)}\right),
\end{equation}
\vspace{-4mm}
\begin{equation}
	\pi\left(s_{t}, \widetilde{\mathcal{S}}_{(t, t+1)}\right) = \mathbf{F}_{\textrm{in}}\left[\ \arg\max(\textrm{M}_{\textrm{agent}})\ \right],
\end{equation}
where $\parallel$ denotes the concatenation operation and $\mathbf{F}_{\textrm{in}}$ refers to the input feature map.

However, the weights of $\textrm{MLP}_{\textrm{agent}}$ cannot be updated smoothly during backpropagation due to the undifferentiable $\arg\max$ function. To solve this problem, we use a Gumbel Softmax function~\cite{jang2016categorical,yang2019modeling}, which is computed as a one-hot vector for forward operation, and updates the weight using the results of the softmax function for backward propagation. With these methods, our curves are represented as follows:
\begin{equation}
	\mathbf{curve} = \left[s_{1}\rightarrow s_{2}\rightarrow \cdots\rightarrow s_{T}\right] \in \mathbb{R}^{C \times (T-1)}.
\end{equation}
We set all joint nodes in the first frame to be the starting points of the curves, so the shape of the integrated curve is $\mathbf{curves} \in \mathbb{R}^{C \times(T-1) \times V}$, where $V$ stands for the number of joint nodes.

\paragraph{\bf{Curve Aggregation.}}
Inspired by~\cite{Xiang_2021_ICCV}, we use an aggregation module to effectively apply the curves to the input feature map. The process of the curve aggregation module is shown in~\cref{fig:curves} (b). By applying the curve aggregation, the model can consider both the relationship between nodes existing in one curve (intra-curve features $\mathbf{F}_{intra}$) and the relationship between every curves (inter-curve features $\mathbf{F}_{inter}$). To construct $\mathbf{F}_{intra}$, we first obtain $\mathbf{curve}_{intra}\in\mathbb{R}^{C_{mid}\times (T-1)}$ through the attentive pooling layer~\cite{hu2021learning} and a simple MLP layer that reduces the number of channels. The $\mathbf{F}_{intra}$ is computed as follows:
\begin{equation}
	\label{eq:finter1}
	\widetilde{\mathbf{F}}_{intra} = softmax\left(\mathbf{F}_{\textrm{in}} \times \mathbf{curve}_{intra}\right), 
\end{equation}
\begin{equation}
	\label{eq:finter2}
	\mathbf{F}_{intra} = \mathbf{curve}_{intra}\mathbf{W}_{intra}\times \widetilde{\mathbf{F}}_{intra},
\end{equation}
where $\mathbf{W}_{intra}\in\mathbb{R}^{C_{mid}\times C_{mid}}$ is an MLP layer that linearly transforms the curve features. The $\mathbf{curve}_{intra}$ is applied to the input feature map $\mathbf{F}_{\textrm{in}}$ while the existing feature map shape $\in \mathbb{R}^{C_{mid}\times T\times V}$ is preserved. The $\mathbf{F}_{inter}$ is obtained in a same way to~\cref{eq:finter1} and~\cref{eq:finter2}. Our curve aggregation module is shown in~\cref{fig:curves} (b). To aggregate the two different features, we use the following method:
\begin{equation}
	\mathbf{F}_{\textrm{out}} = \left(\mathbf{F}_{intra} \parallel \mathbf{F}_{inter}\right)\mathbf{W}_{agg} \in \mathbb{R}^{C\times T\times V},
\end{equation}
where $\mathbf{W}_{agg} \in \mathbb{R}^{2C_{mid}\times C}$ integrates $\mathbf{F}_{intra}$ and $\mathbf{F}_{inter}$. Finally, a new feature map $\mathbf{F}_{\textrm{out}}$ is generated, where the curve features are applied to the input feature map $\mathbf{F}_{\textrm{in}}$, and it enables the model to consider the spatio-temporal dependencies.

\begin{figure}[t]
	\centering
	\includegraphics[width=\linewidth]{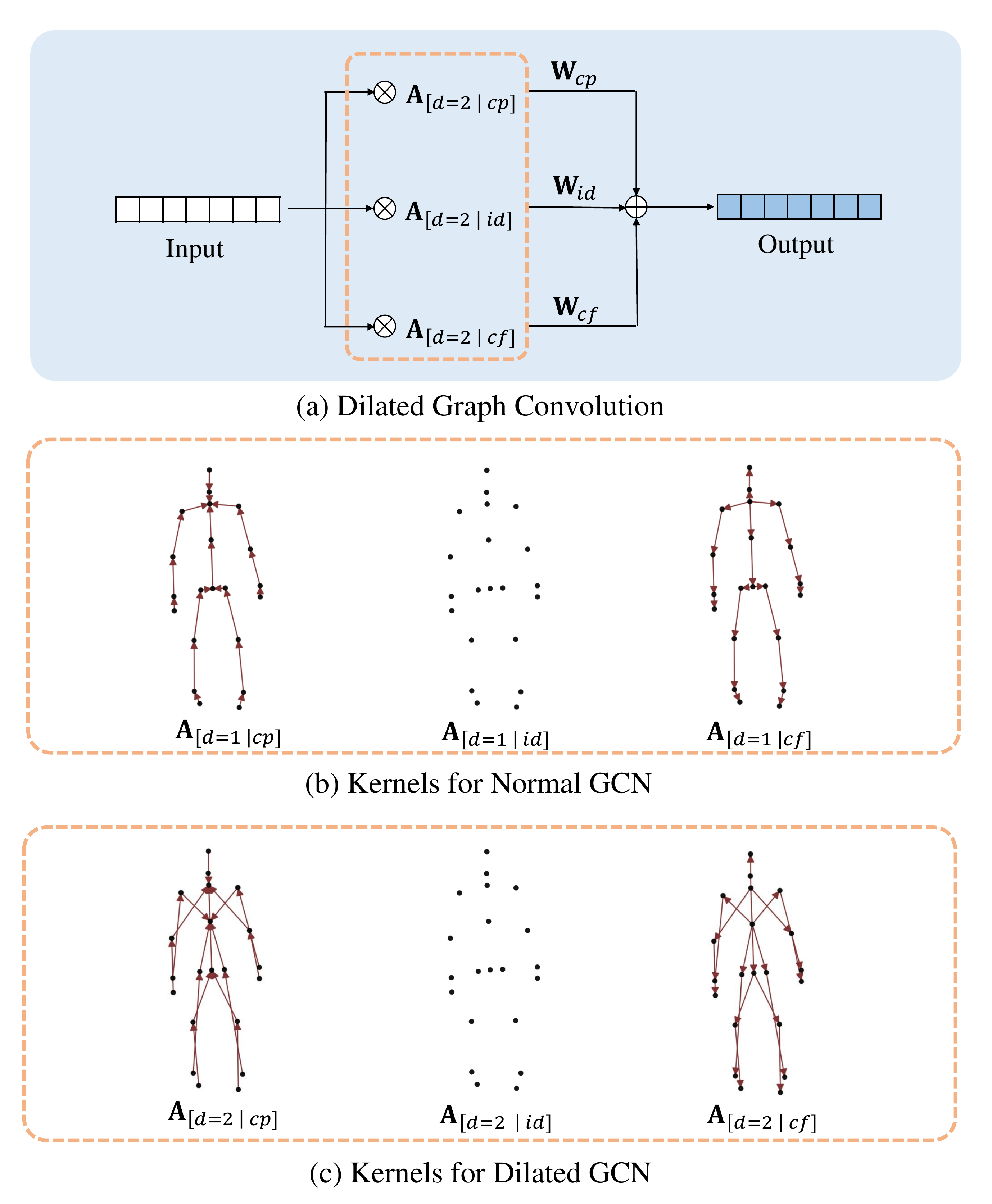}
	\vspace{-5mm}
	\caption{(a) Structure of dilated graph convolution, and comparison of (b) kernels for normal GCN and (c) kernels for dilated GCN. The arrows in (b) and (c) include information about the direction of the edges and the start and end points. The dilation value for kernels of (c) is fixed to 2.}
	\vspace{-3mm}
	\label{fig:dgc}
\end{figure}

\subsection{Dilated Kernels for Graph Convolution}

In this subsection, we propose dilated kernels for graph convolution on the skeletal graph structure, which increase spatial receptive field without any additional parameters.

\paragraph{\bf{Kernel Analogy from CNNs to GCNs.}}

The kernels for CNNs are applied to networks to aggregate local features in pixel units. In particular, for convolution to a single axis, local features are largely divided into [``Left'', ``Identity'', ``Right''] when the size of the kernel is 3. Those features are integrated into a representative information that identifies adjacent pixels by weighted summation. For example, if the ``Left'', ``Identity'', and ``Right'' features are symbolized as -1, 0, and 1, respectively, a kernel with a dilation of 2 can be expressed as [-2, 0, 2]. Therefore, the CNN operation with a dilation of $d$ is as follows:
\begin{equation}
	\mathbf{F}_{\textrm{out}} = \sum_{k \in [-1, 0, 1]}\mathbf{F}_{\textrm{in}\ \left[{\ p\  +\  k\  *\  d\ }\right]}\ \mathbf{W}_{k},
\end{equation}
where $p$ denotes location of the pixel and $\mathbf{W}_{k}$ denotes the weights of the kernel-wise MLP layer.

In non-Euclidean geometry, the concept of ``Right'' and ``Left'' for CNNs cannot be used because those directions cannot be defined, especially for the graph structures. Yan~\etal~\cite{yan2018spatial} first propose a method for selecting the root node and dividing the kernel into [``Centripetal'' ($cp$), ``Identity'' ($id$), ``Centrifugal'' ($cf$)] to aggregate the local features for skeleton-based action recognition. According to this policy, the dilated GCN operation is as follows:
\begin{equation}
	\label{eq:gcn}
	\Phi\left(\mathbf{F}_{\textrm{in}}, d\right) = \sum_{k \in [cp, id, cf]}\mathbf{A}_{\left[d\mid k\right]}\mathbf{F}_{\textrm{in}}\mathbf{W}_{k},
\end{equation}
where $\mathbf{A}_{\left[d\mid k\right]}$ denotes a normalized adjacency matrix according to the dilation window given the direction of the kernel, and $\Phi$ refers to the DK-GC operation. For example, the GCNs with a dilation of 2 aggregate node features by skipping one adjacent node. Our DK-GC operation is systematically similar to that of the CNN (e.g., number of parameters, floating point operations) as shown in~\cref{fig:dgc} (a). In addition, because the $[id]$ kernel itself denotes an identity matrix, $\mathbf{A}_{[d=n\mid id]}$ is the same matrix as $\mathbf{A}_{[d=1\mid id]}$. In other words, adjacency matrices for a kernel size of 3 and a dilation of 2 are divided into $[\mathbf{A}_{[d=2\mid cp]}$, $\mathbf{A}_{[d\mid id]}$, $\mathbf{A}_{[d=2\mid cf]}]$. The kernels for GCNs with adjacent connectivity and DK-GCs are shown in~\cref{fig:dgc} (b) and (c).

\paragraph{\bf{Graph Convolution with Dilated Kernels.}}

To construct adjacency matrices with the dilated kernels, we simply extend the approach of Liu~\etal~\cite{liu2020disentangling}'s method. The $k$ power of an adjacency matrix includes edges that are structurally $k$ steps away from the query node. However, if only the power of the adjacency matrix is used to reflect nodes that are $k$ steps away, the output matrix includes the paths back to the query node. To exclude these paths, we use the difference between the $d$ power and the $(d-1)$ power of the adjacency matrix:
\begin{equation}
	\small
	\begin{aligned}
	\tilde{\mathbf{A}}_{\left[\textrm{d}\mid k\right]} &= \lambda\left((\tilde{\mathbf{A}}_{\left[d\textrm{=}1\mid k\right]}+\mathbf{I})(\tilde{\mathbf{A}}_{\left[d\textrm{=}1\mid cf,cp\right]}+\mathbf{I})^{\textrm{d}-1}\right) \\ &-\lambda\left((\tilde{\mathbf{A}}_{\left[d\textrm{=}1\mid k\right]}+\mathbf{I})(\tilde{\mathbf{A}}_{\left[d\textrm{=}1\mid cf,cp\right]}+\mathbf{I})^{\textrm{d}-2}\right),
	\end{aligned}
\end{equation}
\begin{equation}
	\mathbf{A}_{\left[\textrm{d}\mid k\right]} = \mathbf{D}_{\left[\textrm{d}\mid k\right]}^{-\frac{1}{2}}\tilde{\mathbf{A}}_{\left[\textrm{d}\mid k\right]}\mathbf{D}_{\left[\textrm{d}\mid k\right]}^{-\frac{1}{2}},
\end{equation}
where $\tilde{\mathbf{A}}$ and $\mathbf{A}$ stand for the unnormalized and normalized adjacency matrices, respectively, and $\mathbf{D}$ denotes the degree matrix for normalization. The $\lambda$ function receives input as matrix and replaces all the elements greater than 1 with 1 and all values less than 1 with 0. Without the $\lambda$ function, it hinders the optimization and convergence of the model since the elements for the overlapping paths become largely biased values. The $\lambda$ function allows the model to converge stably while avoiding those biases on the edges for overlapping paths.

\begin{figure}[t]
	\centering
	\includegraphics[width=\linewidth]{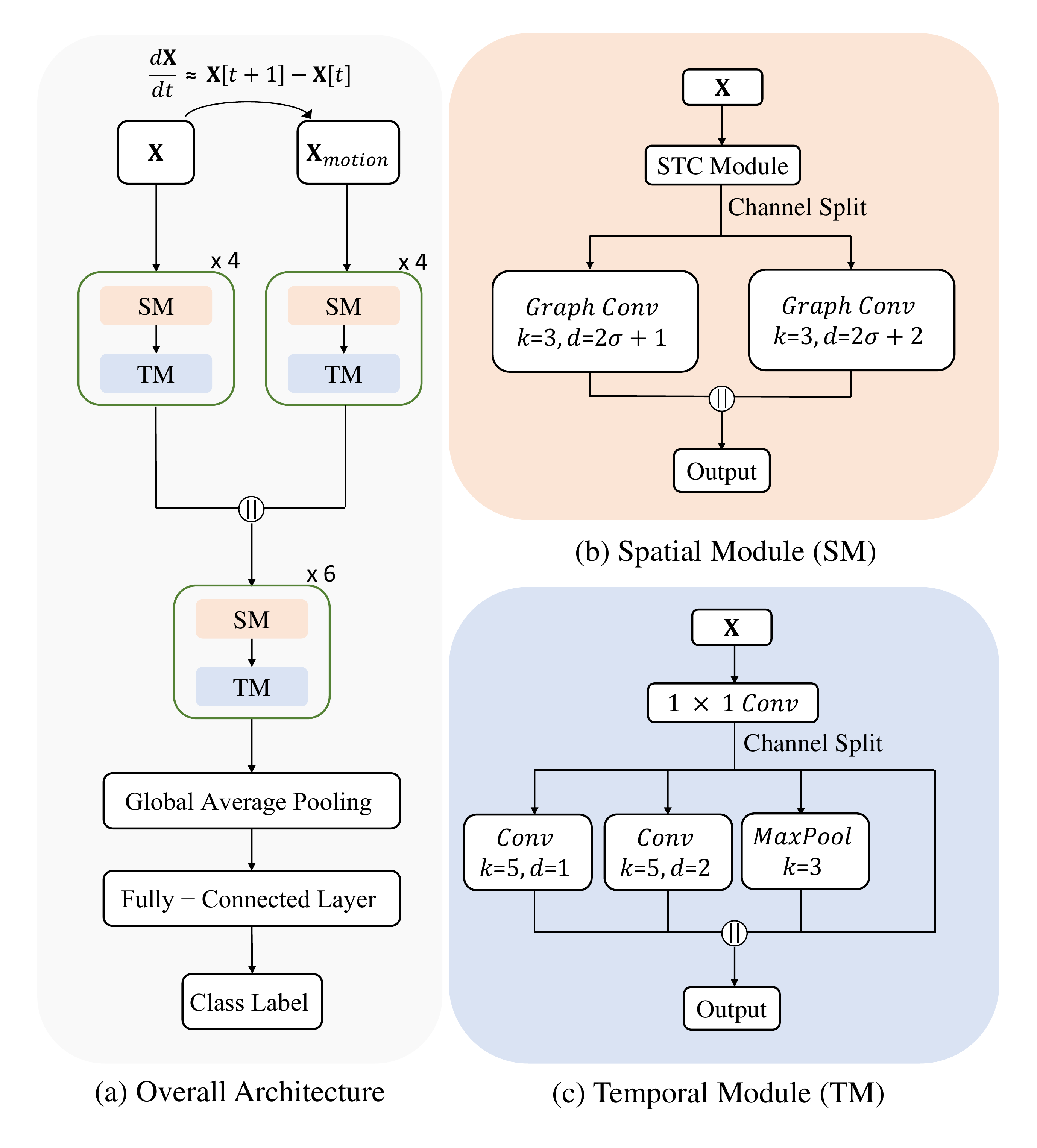}
	\caption{Overall architecture of the model (a) and the spatial (b) and temporal modules (c). Graph convolutions of the spatial module and convolutions of the temporal module are the operations for the node axis and the frame axis, respectively.}
	\label{fig:architecture}
\end{figure}

\vspace{+5mm}
\subsection{Network Architecture} 
\label{sec:network}

\paragraph{\bf{Overall Architecture.}}
We adopt the architecture of~\cite{yan2018spatial} as our baseline, which includes 10 spatio-temporal blocks. The output channels of each block are 64, 64, 64, 64, 128, 128, 128, 256, 256, and 256, and each block includes independent spatial and temporal module, and residual connections~\cite{he2016deep} for stable learning.
For data composed of the time-series $x_{t}$, $y_{t}$, and $z_{t}$ coordinates, we utilize the motion vector $\dfrac{d\mathbf{X}} {dt}\approx [\ x_{t+1}-x_{t},\ y_{t+1}-y_{t},\ z_{t+1}-z_{t}\ ]$ differently from existing methods~\cite{shi2020skeleton,chen2021channel,cheng2020decoupling}. Coordinate and motion data pass through four spatio-temporal blocks independently, and then they are concatenated to take account of those two modalities in a single network. Our overall architecture is shown in~\cref{fig:architecture} (a).

\paragraph{\bf{Spatial Module.}} \label{sec:spat}
We construct a multi-branch spatial module based on DK-GC. Our spatial module has 2 layers in series and 2 operations in parallel, as shown in~\cref{fig:architecture} (b). This module first passes the input feature map $\mathbf{F}_{\textrm{in}}$ through a STC module, which generates the curves and contains curve aggregation module. For efficient computational complexity, we place the STC modules only on specific layers of the model, and place point-wise convolution on the remaining layers. After the block, we split the transformed features into two branch operations along the channel axis, and feed them into two branch operations with an dilation scaling factor $\sigma$ and fixed kernel size 3:
\begin{equation}
	\mathbf{F}_{\textrm{out}} = \concat_{\tilde{d} \in [1,2]}\Phi\left(\mathbf{F}_{\textrm{in}},\ d=2\sigma + \tilde{d}\right),
\end{equation}
where $\Phi$ denotes the DK-GC operation in~\cref{eq:gcn} and $d$ denotes the dilation of each GCN operation. Then, the two branch operation results are concatenated to create a new output feature map $\mathbf{F}_{\textrm{out}}$. Here, we train the model separately for $\sigma\in\{0, 1, 2\}$. The trained models are then used for ensemble, which enables us to combine the strengths of each individual model and improve overall performance.

\paragraph{\bf{Temporal Module.}}
We adopt the multi-scale temporal convolution of~\cite{chen2021channel} as the temporal module shown in~\cref{fig:architecture} (c). This module includes one feature transformation block and four branch operations, similar to the spatial module. Two of their operations are temporal convolutions with dilations of 1 and 2, which have a kernel size of 5. The remaining operations are a max pooling layer with a kernel size of 3 and a identity layer. After all four operations are completed, the output feature map is constructed by concatenating all the resulting feature maps.
\begin{table*}[]{
		\begin{center}
						\resizebox{\textwidth}{!}{
					\begin{tabular}{l||c|c|c|c|c|c|c|c}
						\toprule
						\multirow{2}{*}{Methods}  & \multirow{2}{*}{Publication} & \multicolumn{2}{c|}{NTU-RGB+D 60} & \multicolumn{2}{c|}{NTU-RGB+D 120} & \multicolumn{2}{c|}{Kinetics-Skeleton}  & Northwestern\\ \cmidrule{3-8}
						\multicolumn{1}{c||}{}            &              & X-Sub (\%)      & X-View (\%)  & X-Sub (\%) & X-Set (\%) & Top-1 (\%) & Top-5 (\%) & UCLA    \\ \midrule \midrule
						ST-GCN~\cite{yan2018spatial}       & AAAI 2018                                & 81.5            & 88.3      & 82.5            & 84.2      & 30.7 & 52.8 & - \\
						2s-AGCN~\cite{shi2019two}           &  CVPR 2019                        & 88.5            & 95.1      & 88.5            & 95.1       & 36.1 & 58.7& - \\
						DGNN~\cite{shi2019skeleton}            & CVPR 2019                            & 89.9            & 96.1     & - & -      & 36.9 & 59.6 & - \\
						AGC-LSTM~\cite{si2019attention}         & CVPR 2019                               & 89.2            & 95.0  & -            & - & - & - & 93.3        \\
						Shift-GCN~\cite{cheng2020skeleton}      & CVPR 2020                             & 90.7            & 96.5   & 85.9            & 87.6 & - & - & 94.6       \\
						DC-GCN+ADG~\cite{cheng2020decoupling}    & ECCV 2020                              & 90.8            & 96.6    & 86.5            & 88.1   & - & - & 95.3     \\
						MS-G3D~\cite{liu2020disentangling}       & CVPR 2020                              & 91.5            & 96.2   & 86.9            & 88.4         & 38.0 & 60.9 & - \\
						MST-GCN~\cite{chen2021multi}             & AAAI 2021                      & 91.5            & 96.6   & 87.5            & 88.8      & 38.1 & 60.8 & - \\
						DDGCN~\cite{korban2020ddgcn}           & ECCV 2020 & 91.1         	 & 97.1 	  		& -            & -    & 38.1 & 60.8 & -\\         
						CTR-GCN~\cite{chen2021channel}              & ICCV 2021                        & 92.4            & 96.8        & 88.9            & 90.6     & - & - & 96.5 \\
						EfficientGCN-B4~\cite{song2022constructing}  & TPAMI 2022                            & 91.7            & 95.7     & 88.3            & 89.1         & - & - & - \\
						STF~\cite{ke2022towards}                      & AAAI 2022                   & 92.5            & 96.9         & 88.9            & 89.9     & 39.9 & - & - \\
						InfoGCN (4-ensemble)~\cite{chi2022infogcn}  		                 & CVPR 2022      & 92.7            & 96.9     & 89.4            & 90.7     & - & - & 96.6\\ 
						InfoGCN (4-ensemble)~\cite{chi2022infogcn}  		                 & CVPR 2022      & \cellcolor{yellow}{93.0}            & \cellcolor{yellow}{97.1}     & 89.8            & 91.2     & - & -& 97.0\\ \midrule
						STC-Net (2-ensemble)			& 		&	92.5 & 96.7	& 89.3	& 90.7 & 40.0 & 62.6 & 96.8 \\
						STC-Net (4-ensemble)                                      &   & \cellcolor{yellow}{93.0}   & \cellcolor{yellow}{97.1} & \cellcolor{yellow}{89.9}   & \cellcolor{yellow}{91.3} & \cellcolor{yellow}{40.7} & \cellcolor{yellow}{63.6} & \cellcolor{yellow}{97.2} \\ 
						STC-Net (6-ensemble)                                      &   & \cellcolor{orange}{93.3}   & \cellcolor{orange}{97.3} & \cellcolor{orange}{90.2}   & \cellcolor{orange}{91.7}& \cellcolor{orange}{41.2} & \cellcolor{orange}{64.2} & \cellcolor{orange}{97.4} \\ \bottomrule
					\end{tabular}
			}
		\end{center}
		\vspace{-2mm}
		\caption{\textbf{Comparison of the top-1 (or 5) accuracy (\%) with the state-of-the-arts on NTU-RGB+D 60, NTU-RGB+D 120, Kinetics-Skeleton, and Northwestern-UCLA datasets.} The \colorbox{orange}{orange} and \colorbox{yellow}{yellow} cells respectively indicate the highest and second-highest value.}
		\vspace{-4mm}
		\label{tab:ntu}
	}
\end{table*}

\section{Experiments}
\label{sec:exp}

\subsection{Datasets}

\paragraph{\bf{NTU-RGB+D 60.}}
NTU-RGB+D 60~\cite{shahroudy2016ntu} is a large skeleton-based action recognition dataset that contains 56,880 action sequences. All the sequences are classified into a total of 60 classes. This dataset is captured by three Microsoft Kinect v2 cameras, each of which captures at three horizontal angles. At most two subjects exist in an action sample. We follow two benchmarks suggested by the authors of the dataset. (1) Cross-Subject (X-Sub): The actions of 20 out of 40 subjects are used for training, and the actions of the remaining 20 are used for validation. (2) Cross-View (X-View): Two of the three camera views are used for training, and the other one is used for validation.

\paragraph{\bf{NTU-RGB+D 120.}}

NTU-RGB+D 120~\cite{liu2019ntu} is a dataset in which 57,367 action sequences are added from the NTU-RGB+D 60 dataset. All action sequences are classified into a total of 120 classes. All the samples are captured with three camera views. In addition, the dataset is captured under 32 settings. We use two benchmarks proposed by the authors of this dataset. (1) Cross-Subject (X-Sub): The actions of 53 objects out of 106 objects are used for training, and the rest are used for validation. (2) Cross-Setup (X-Set): Among 32 numbered settings, even-numbered settings are used for training, and odd-numbered settings are used for validation.

\paragraph{\bf{Kinetics-Skeleton.}} 

The Kinetics-Skeleton dataset is derived from the Kinetics 400 video dataset~\cite{kay2017kinetics}, utilizing the OpenPose pose estimation~\cite{cao2017realtime} to extract 240,436 training and 19,796 testing skeleton sequences across 400 classes. Each skeleton graph includes 18 body joints with their 2D spatial coordinates, and prediction confidence scores from OpenPose. We report Top-1 and Top-5 accuracies, following the convention.

\paragraph{\bf{Northwestern-UCLA.}}

Northwestern-UCLA~\cite{wang2014cross} is an action recognition dataset containing 1494 skeleton sequences. All the action samples are classified into 10 classes that are captured by three cameras at different angles. We use the protocol proposed by the authors: two of the three camera views are used for training and the other is for validation.

\subsection{Experimental Settings}

In our experiments, we set the number of epochs to 90, and we apply a warm-up strategy~\cite{he2016deep} to the first five epochs for more stable learning. We adopt an SGD optimizer with a Nesterov momentum of 0.9 and a weight decay of 0.0004. The initial learning rate is set to 0.1, and we reduce the learning rate to 0.0001 through the cosine annealing scheduler~\cite{loshchilov2016sgdr}. We use Zhang~\etal~\cite{zhang2020semantics}'s data preprocessing method for NTU-RGB+D 60 and 120 datasets, and we set the batch size to 64. For the Northwestern-UCLA dataset, we use the data preprocessing method by Cheng~\etal~\cite{cheng2020skeleton} and set the batch size to 16. For the Kinetics-Skeleton, we set the batch size to 64. In addition, the STC module is applied to the 3-rd, 6-th, and 9-th blocks of the 10 spatio-temporal blocks for memory efficiency, and pointwise convolution is applied to the remaining blocks. Our experiments are conducted on a single RTX 3090 GPU.

\subsection{Comparison with the State-of-the-Arts}
Many recent state-of-the-art models~\cite{shi2020skeleton,chen2021channel,cheng2020decoupling} use the ensemble method by training four data streams, i.e., joint, bone, joint motion, and bone motion. However, as networks with only the joint (bone) motion stream show inferior performances than networks with joint (bone) stream, learning two streams independently is inefficient. Unlike them, we train joint and joint motion streams on one network, as shown in~\cref{fig:architecture} (a). We also train bone and bone motion streams on the network. We adopt the ensemble method of models trained with the dilation scaling factor $\sigma$ as 0, 1, and 2 without considering the motion stream separately, where $\sigma$ is explained in~\cref{sec:spat}. In other words, we use the six models with the joint ($\sigma\in\{0, 1, 2\}$), bone ($\sigma\in\{0, 1, 2\}$) streams for our ensemble method.

We evaluate performance on four skeleton-based action recognition benchmarks. The performance comparisons for those datasets are shown in \cref{tab:ntu}. We systematically evaluate the performance of our STC-Net with respect to the number of ensemble streams, and show that the 6-stream ensemble outperforms state-of-the-art performance on all datasets. For the all datasets, even with 4 ensembles, the STC-Net shows slightly better or about the same as~\cite{chi2022infogcn}, a state-of-the-art model with 6-stream ensemble.
\begin{table}[]
	\begin{center}
		\resizebox{\columnwidth}{!}{
			\begin{tabular}{c|c|ccc|c|cc}
				\toprule
				\multirow{2}{*}{Methods} & \multirow{2}{*}{M} & \multicolumn{2}{c}{Curve} 	& \multirow{2}{*}{CA}       & \multirow{2}{*}{DK-GC}		& \multirow{2}{*}{X-Sub (\%)} & \multirow{2}{*}{X-Set (\%)} \\ \cmidrule{3-4}
				&                     & Intra         & Inter       & 							&  								&							  &                            \\ \midrule \midrule
				A		            &         				  & 			  &             &               			&    							& 83.5                            & 85.4                            \\
				B                   & \ding{51}               & 			  &        		&               			&           					& 84.8 ($\uparrow$ 1.3)           & 86.6 ($\uparrow$ 1.2)            \\ \midrule
				C                   & \ding{51}               & \ding{51}	  &           	&               			&          						& 85.2 ($\uparrow$ 1.7)           & 86.9 ($\uparrow$ 1.5)                           \\
				D                   & \ding{51}               & 			  & \ding{51}   &              				&           					& 85.1 ($\uparrow$ 1.6)           & 86.9 ($\uparrow$ 1.5)                           \\
				E                   & \ding{51}               & 			  &    			&              				& \ding{51}     				& 85.8 ($\uparrow$ 2.3)           & 87.3 ($\uparrow$ 1.9)                           \\ 
				F                   & \ding{51}               & \ding{51}	  & \ding{51}   &              				&           					& 85.4 ($\uparrow$ 1.9)           & 87.2 ($\uparrow$ 1.8)                           \\ \midrule
				G                   & \ding{51}               & \ding{51}	  & \ding{51}   &              				& \ding{51}     				& 86.0 ($\uparrow$ 2.5)           & 87.7 ($\uparrow$ 2.3)                           \\ 
				H                   & \ding{51}               & \ding{51}	  & \ding{51}  	& \ding{51}             	&           					& 85.9 ($\uparrow$ 2.4)           & 87.7 ($\uparrow$ 2.3)                           \\
				I                   & \ding{51}               & \ding{51}	  & \ding{51}  	& \ding{51}             	& \ding{51}     				& \textbf{86.2} ($\uparrow$ 2.7)  & \textbf{88.0} ($\uparrow$ 2.6)                           \\ \bottomrule
			\end{tabular}
		}
	\end{center}
	\vspace{-2mm}
	\caption{\textbf{Performance comparison of variants of STC-Net.} M: motion, Intra: $\mathbf{curve}_{intra}$, Inter: $\mathbf{curve}_{inter}$, CA: Curve Aggregation module}
	\vspace{-6mm}
	\label{tab:ablifc}
\end{table}

\subsection{Ablation Study}
In this section, we conduct several experiments to prove the superiority of our proposed modules. The experiments are performed by dividing the components of our STC-Net to $\mathbf{curve}_{intra}$, $\mathbf{curve}_{inter}$, curve aggregation module, and DK-GC. The performance described in this section refers to cross-subject and cross-setup accuracy on the NTU-RGB+D 120 joint stream.

\begin{figure}[t]
	\centering
	\includegraphics[width=\linewidth]{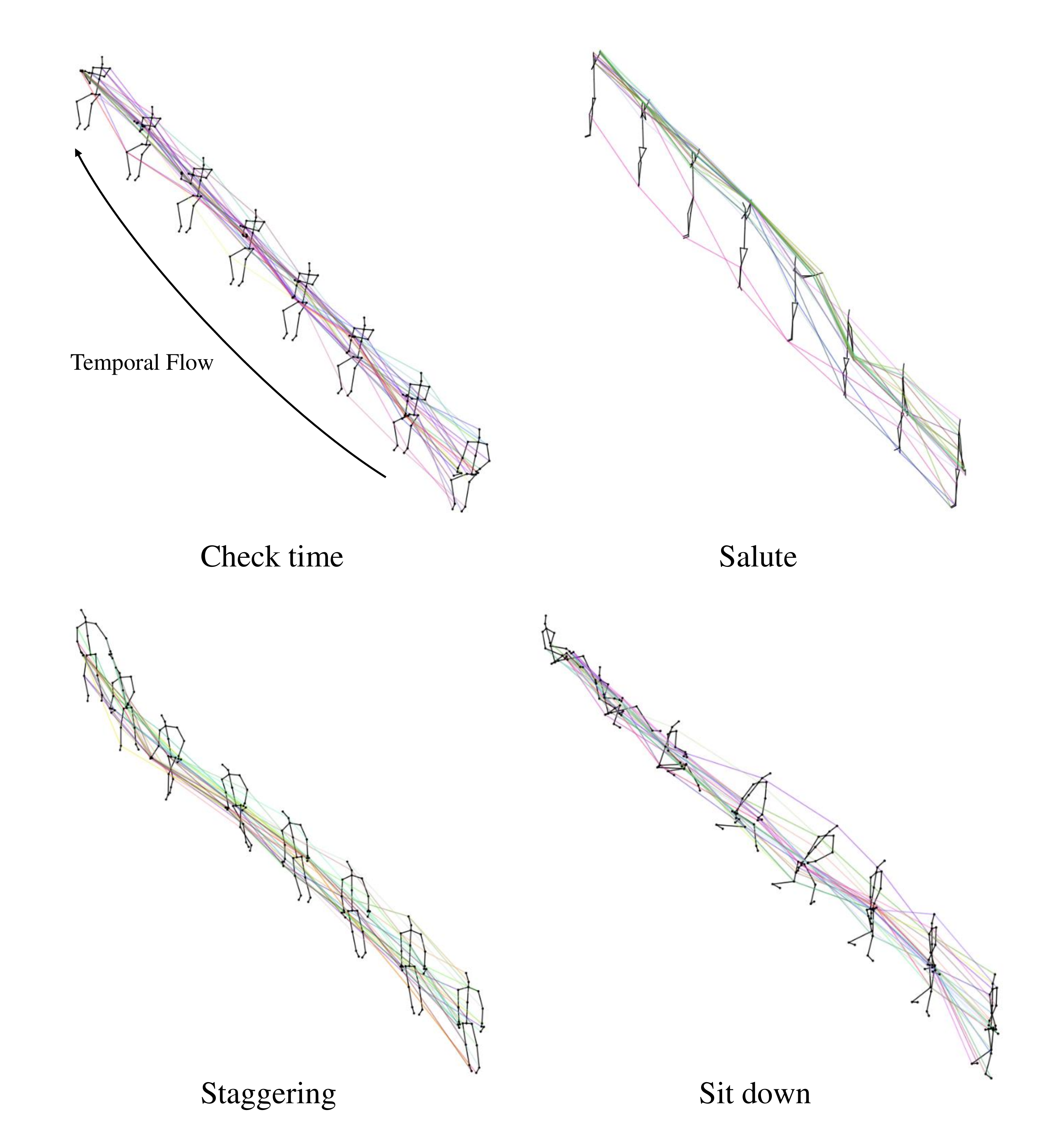}
	\vspace{-4mm}
	\caption{Curve visualizations for several samples. To make a clear distinction, each curve is represented by a unique color.}
	\label{fig:vis}
	\vspace{-5mm}
\end{figure}

\paragraph{Spatio-Temporal Curve Module.}
In order to prove the effectiveness of the STC module, we specify a baseline model in which the temporal module of~\cite{chen2021channel} is applied to the architecture in~\cite{yan2018spatial}. According to~\cref{tab:ablifc}, the performance increases by 1.25\% only with motion data. If we use either $\mathbf{curve}_{intra}$ or $\mathbf{curve}_{inter}$ via simple non-local block structure~\cite{wang2018non} instead of curve aggregation module, the performance improves slightly. It is due to the failure to properly utilize the two curve features designed for attention to the input feature map. Thus, we observe that the effect is more pronounced when using the curve aggregation module. The model H in~\cref{tab:ablifc} includes all the components of the spatio-temporal curve and shows a performance that is 2.35\% higher than the baseline model A.

\vspace{-2mm}
\begin{table}[h]
	{
		\begin{center}
			\resizebox{\columnwidth}{!}{
				\begin{tabular}{l|c|cc|cc}
					\toprule
					Methods								& Kernel Size						& X-Sub (\%)		& X-Set (\%)		& Params					& FLOPs\\ \midrule \midrule
					STC-Net 							& 3									& 86.2			    & 88.0			    & 1.46 M          			& 1.88 G          \\
					STC-Net 							& 5									& 86.4			    & 88.1			    & 1.78 M          			& 2.32 G          \\
					STC-Net								& 7									& 86.2				& 87.9				& 2.09 M					& 2.76 G		\\ \bottomrule
				\end{tabular}
			}
		\end{center}
		\vspace{-3mm}
		\caption{Comparison of different DK-GC based on kernel size.}
		\vspace{-6mm}
		\label{tab:abldgc}
	}
\end{table}

\begin{table}[t]
	{
		\begin{center}
			\resizebox{\columnwidth}{!}{
				\begin{tabular}{l|c|cc|cc}
					\toprule
					\multirow{2}{*}{Methods}			& \multirow{2}{*}{\# Ensembles}		& \multicolumn{2}{c|}{NTU-RGB+D 120}	& \multirow{2}{*}{Params}	& \multirow{2}{*}{FLOPs}	\\ \cmidrule{3-4}
					& 									& X-Sub (\%) 		& X-Set (\%)    	& 							& \\ \midrule \midrule
					MS-G3D~\cite{liu2020disentangling} 	& 2									& 86.9			    & 88.4			    & 6.44 M          			& 24.50 G          \\					
					InfoGCN~\cite{chi2022infogcn}       & 2									& 88.5              & 89.7          	& 3.14 M          			& \textbf{3.36 G}          \\					
					\textbf{STC-Net} 					& 2			 						& \textbf{89.3}		& \textbf{90.7}		& \textbf{2.92 M}			& 3.70 G \\  \midrule
					DC-GCN~\cite{cheng2020decoupling} 	& 4									& 86.5			    & 88.1			    & 13.80 M          			& 51.52 G         \\
					MST-GCN~\cite{chen2021multi}		& 4									& 87.5				& 88.8				& 11.68 M					& 67.12 G		\\
					CTR-GCN~\cite{chen2021channel}      & 4									& 88.9              & 90.6			    & 5.84 M          			& 7.88 G \\
					InfoGCN~\cite{chi2022infogcn}       & 4									& 89.4              & 90.7          	& 6.28 M          			& \textbf{6.72 G}          \\
					\textbf{STC-Net} 					& 4 								& \textbf{89.9}		& \textbf{91.3}		& \textbf{5.84 M}			& 7.40 G \\  \midrule
					InfoGCN~\cite{chi2022infogcn}       & 6									& 89.8              & 91.2          	& 9.42 M          			& \textbf{10.08 G}          \\
					\textbf{STC-Net} 		 			& 6 								& \textbf{90.2}		& \textbf{91.7}		& \textbf{8.76 M}			& 11.10 G\\ \bottomrule 
				\end{tabular}
			}
		\end{center}
		\vspace{-2mm}
		\caption{Comparison of multi-stream complexity of the state-of-the-arts according to the number of ensemble streams.}
		\vspace{-5mm}
		\label{tab:complexity}
	}
\end{table}

\paragraph{Dilated Kernels for Graph Convolution.}
According to~\cref{tab:ablifc}, the model E with only DK-GC shows 2.1\% higher performance than the baseline model A. Furthermore, its effectiveness is more pronounced when used with the STC module (model I), which is 2.65\% higher than the baseline model A. To find the optimal kernel size of the DK-GC, we conduct additional experiments by setting the kernel size to 3, 5, and 7 with the dilation scaling factor $\sigma = 0$.~\cref{tab:abldgc} shows that performance of our model does not change according to the size of the kernel. It is due to the relatively small number of joint nodes for action recognition. Therefore, considering that there is little difference in performance, selecting the smallest kernel size of 3 is most efficient in term of parameters and computation.

\vspace{-1mm}
\subsection{Curve Visualization}
To qualitatively evaluate whether the curves are well-generated, we visualize the curves for several samples as shown in~\cref{fig:vis}. For the ``Check time'' and ``Salute'' classes, the curves start from every node in the first frame, and those curves tend to proceed toward hand or arm nodes. Inspired by human visual recognition, it is reasonable that the hand gestures should be highlighted for those classes. Similarly, the curves of ``Staggering'' and ``Sit down'' classes tend to directed toward lower body, which is also reasonable in that the leg gestures are important for those classes.

\vspace{-1mm}
\subsection{Analysis of Computational Complexity}
A comparison of the complexity between our model and state-of-the-arts is shown in~\cref{tab:complexity}. Our 4-stream STC-Net shows slightly better performance than 6-stream InfoGCN~\cite{chi2022infogcn} while having the fewer GFLOPs ($\times$0.62) and fewer parameters ($\times$0.73). With 6-stream ensemble, our model outperforms~\cite{chi2022infogcn} by a larger margin.

\vspace{-1mm}
\section{Conclusions}
\label{sec:conclusion}
In this paper, we propose a novel Spatio-Temporal Curve Network (STC-Net) for skeleton-based action recognition, which consists of spatial modules with an spatio-temporal curve (STC) module and graph convolution with dilated kernels (DK-GC). Our STC module constructs spatio-temporal curves by connecting the most highly correlated nodes in successive frames, which significantly increases the spatio-temporal receptive field. Our DK-GC is carefully designed for skeleton-based action recognition to give the model a large spatial receptive field through dilated graph kernels. By combining these two methods, we implement STC-Net and demonstrate its superiority through extensive experiments, and our proposed model outperforms existing methods on four benchmarks.
\clearpage

\appendix
\begin{center}{\bf \Large Appendix}\end{center}
\section{Additional Experimental Results}
Every experimental result of NTU-RGB+D 60~\cite{shahroudy2016ntu}, 120~\cite{liu2019ntu}, Kinetics-Skeleton~\cite{kay2017kinetics} and Northwestern-UCLA~\cite{wang2014cross} datasets are shown in~\cref{tab:appendix2} and~\cref{tab:appendix3}. By applying our ensemble method, each individual model with different dilation scaling factor $\sigma$s and streams contributes to achieving the best performance.

\begin{table}[h]
	
	\resizebox{\columnwidth}{!}{
		\begin{tabular}{c|c|cc|cc}
			\toprule
			\multirow{2}{*}{Stream}   & \multirow{2}{*}{factor $\sigma$} & \multicolumn{2}{c|}{NTU-RGB+D 60} & \multicolumn{2}{c}{NTU-RGB+D 120} \\ \cmidrule{3-6} 
			&                        & X-Sub (\%)    & X-View (\%)   & X-Sub (\%)    & X-Set (\%)    \\ \midrule \midrule
			\multirow{3}{*}{Joint} & 1                   & 91.0              & 96.0          & 86.2             & 87.9           \\
									& 2                    & 90.9              & 96.1          & 86.1             & 88.0           \\ 
									& 3                    & 90.9              & 96.2          & 86.0             & 87.7           \\ \midrule									
			\multirow{3}{*}{Bone} & 1                   & 91.1              & 95.6          & 87.2             & 88.6           \\
									& 2                   	& 91.2              & 95.3          & 87.1             & 88.7           \\ 
									& 3                   	& 91.3              & 95.5          & 87.0             & 88.9           \\ \midrule
			\multicolumn{2}{c|}{Ensemble}                   & \textbf{93.3}              & \textbf{97.3}          & \textbf{90.2}              & \textbf{91.7}          \\ \bottomrule
		\end{tabular}
	}
	\vspace{+2mm}
	\caption{Experimental results of NTU-RGB+D 60 and 120 datasets according to data streams and dilation scaling factor $\sigma$ of our spatial module.}
	\label{tab:appendix2}
	\vspace{-2mm}
\end{table}

\begin{table}[h]
	
	\resizebox{\columnwidth}{!}{
		\begin{tabular}{c|c|cc|c}
			\toprule
			\multirow{2}{*}{Stream}   & \multirow{2}{*}{factor $\sigma$} & \multicolumn{2}{c|}{Kinetics-Skeleton} & Northwestern \\  \cmidrule{3-4}
			&                        & Top-1 (\%)    & Top-5 (\%)   & UCLA (\%)    \\ \midrule \midrule
			\multirow{3}{*}{Joint} & 1                   & 37.9              & 61.0          & 94.8               \\
									& 2                    & 37.5              & 60.8          & 95.7             \\ 
									& 3                    & 37.2              & 60.3          & 95.3             \\ \midrule									
			\multirow{3}{*}{Bone} & 1                   & 37.4              & 60.3          & 95.0             \\
									& 2                   	& 37.5              & 60.5          & 94.0             \\ 
									& 3                   	& 37.2              & 60.1          & 93.8             \\ \midrule
			\multicolumn{2}{c|}{Ensemble}                   & \textbf{41.2}              & \textbf{64.2}          & \textbf{97.4}              \\ \bottomrule
		\end{tabular}
	}
	\vspace{+2mm}
	\caption{Experimental results of Kinetics-Skeleton and Northwestern-UCLA datasets according to data streams and dilation scaling factor $\sigma$ of our spatial module.}
	\label{tab:appendix3}
	\vspace{-2mm}
\end{table}

\begin{figure}[!h]
	\centering
	\includegraphics[width=\columnwidth]{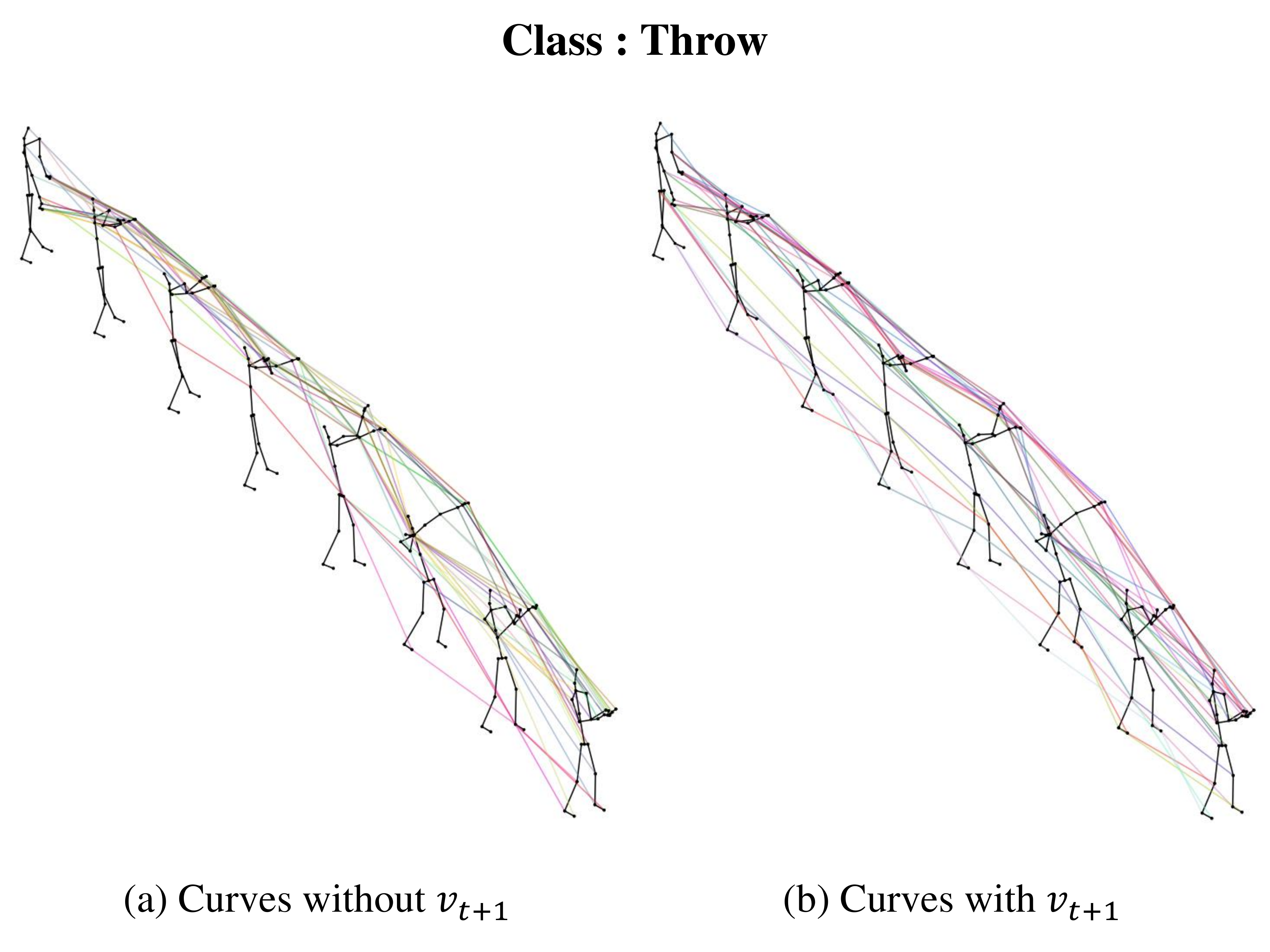}
	\caption{Visualizations for comparing the curves without $v_{t+1}$ and the curves with $v_{t+1}$}
	\label{fig:appendix2}
	\vspace{-2mm}
\end{figure}

\begin{table}[h]
	
	\resizebox{\columnwidth}{!}{
		\begin{tabular}{c|c|ccc|cc}
			\toprule
			\multirow{3}{*}{Method}   & \multirow{3}{*}{DK-GC}	& \multicolumn{3}{c}{Curve Type}									& \multicolumn{2}{|c}{NTU-RGB+D 120} \\ \cmidrule{3-7}
			& 						 & Curve 					& Curve					& Straight  		& \multirow{2}{*}{X-Sub (\%)}	& \multirow{2}{*}{X-Set (\%)} \\
			&                      & w/o $v_{t+1}$		& with $v_{t+1}$		& Line   			&    		&    \\ \midrule \midrule
			Baseline (+ M)				& 		 				&						& 						& 					& 84.8				& 86.6			\\ \midrule
			A							& 		 				&						&						& \ding{51}			& 85.4				& 87.2			\\
			B							& 		 				&						& \ding{51}				& 					& 85.7				& 87.5			\\
			C							& 		 				& \ding{51}				& 						&					& 85.9				& 87.7			\\ \midrule
			D							& \ding{51}		 		& 						& 						&					& 85.8				& 87.3			\\ 
			E							& \ding{51}		 		& 						& 						& \ding{51}			& 85.9				& 87.7			\\
			F							& \ding{51}				&						& \ding{51}			 	& 					& \textbf{86.2}				& 87.8			\\
			G							& \ding{51}				& \ding{51}				& 						&					& \textbf{86.2}		& \textbf{88.0}			\\ \bottomrule
		\end{tabular}
	}
	\vspace{+2mm}
	\caption{Comparison of the STC module variants. M: motion data}
	\label{tab:appendix4}
	\vspace{-2mm}
\end{table}

\begin{figure*}[]
	\centering
	\includegraphics[width=\textwidth]{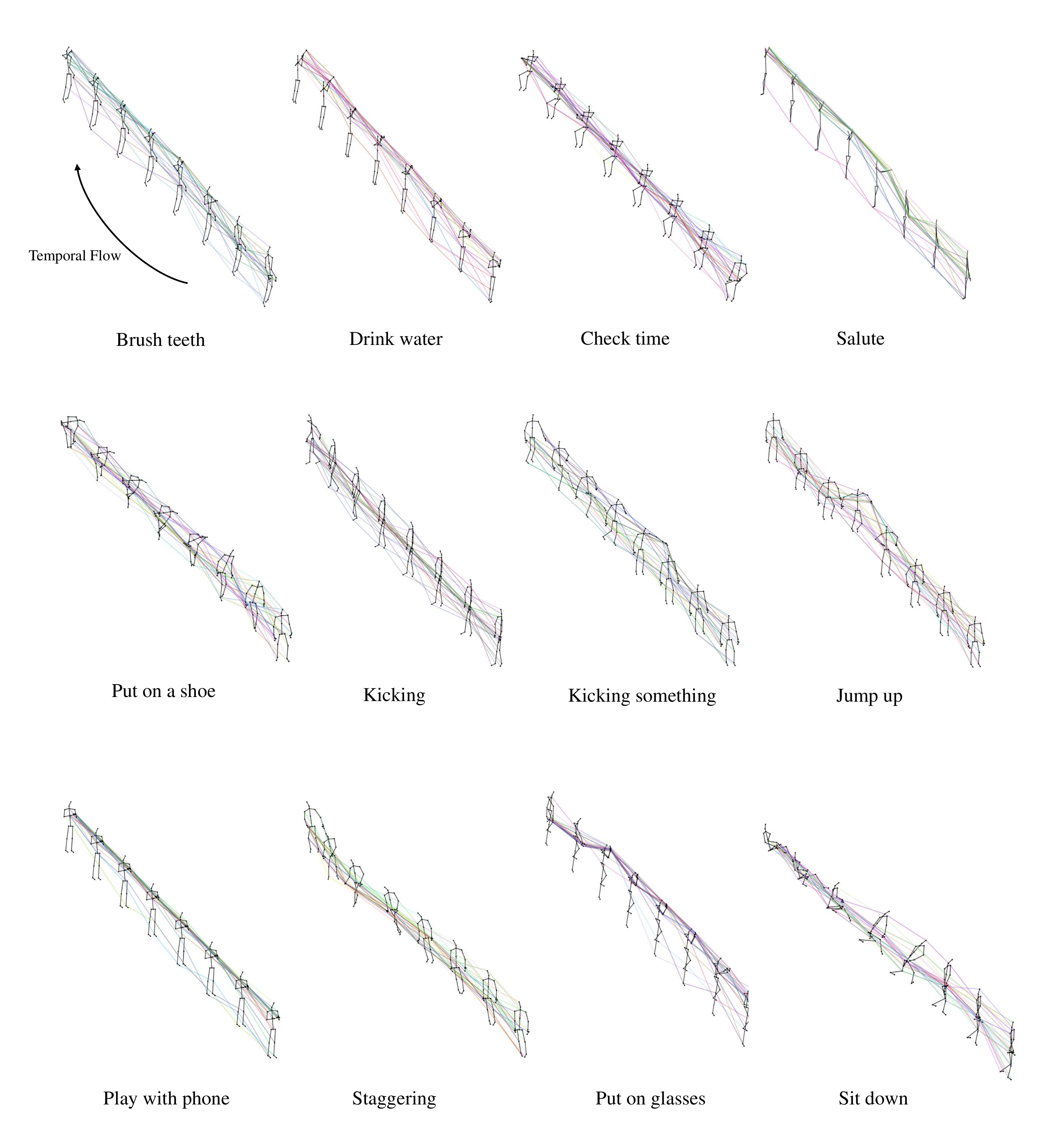}
	\caption{Visualization of trained curves for several data samples.}
	\label{fig:appendix1}
	\vspace{-2mm}
\end{figure*}

\section{Additional Ablation Study}

To verify the superiority of our STC module, we conduct several additional experiments for ablation study. Firstly, we compare the model with our curves to the model with straight lines to prove that increasing spatio-temporal receptive field via STC module is technically effective. The straight lines are implemented by restricting the value of $k$ for k-NN to 1. As shown in~\cref{tab:appendix4}, although the utilization of curves leads to higher performance of the network, even the models A and E that use straight lines outperforms the baseline model as the straight lines utilize all frames at once, resulting in a large temporal receptive field. In addition, we have mentioned in our main paper that choosing structurally identical node $v_{t+1}$ during inter-frame k-NN hinders the model's ability to capture diverse curves. To prove it quantitatively and qualitatively, we compare the performance of the model including curves without $v_{t+1}$ and the model including curves with $v_{t+1}$, and visualize them in~\cref{fig:appendix2} (``throw'' class). As shown in~\cref{tab:appendix4}, the model C and G respectively outperforms the model B and F. Furthermore, referring to~\cref{fig:appendix2} (a) and (b), it can be observed that the curves that do not include $v_{t+1}$ tend to point towards the optimal nodes, namely the hands and arms, whereas the curves that do include $v_{t+1}$ also point towards the hands and arms but also exhibit a tendency to point towards suboptimal nodes (e.g., knees and hips).

\section{Qualitative Results of Inter-Frame Curves}
We propose the Spatio-Temporal Curve (STC) module to identify spatio-temporal dependencies of the human skeleton. Additional qualitative results of the module are shown in~\cref{fig:appendix1}. As mentioned in our main paper, the curves start from every node in the first frame and tend to proceed toward the primary joints for each sequence. Inspired by human visual recognition, it is reasonable to highlight hand gestures for the ``Brush teeth'', ``Drink water'', and ``Salute'' classes, and leg gesture for the ``Put on the shoe'', ``Kicking'', and ``Kicking something'' classes.


%

{\small
\bibliographystyle{ieee_fullname}
\bibliography{egbib}

\begin{thebibliography}{10}\itemsep=-1pt

\bibitem{cao2017realtime}
Zhe Cao, Tomas Simon, Shih-En Wei, and Yaser Sheikh.
\newblock Realtime multi-person 2d pose estimation using part affinity fields.
\newblock In {\em Proceedings of the IEEE conference on computer vision and
  pattern recognition}, pages 7291--7299, 2017.

\bibitem{chen2021channel}
Yuxin Chen, Ziqi Zhang, Chunfeng Yuan, Bing Li, Ying Deng, and Weiming Hu.
\newblock Channel-wise topology refinement graph convolution for skeleton-based
  action recognition.
\newblock In {\em Proceedings of the IEEE/CVF International Conference on
  Computer Vision}, pages 13359--13368, 2021.

\bibitem{chen2021multi}
Zhan Chen, Sicheng Li, Bing Yang, Qinghan Li, and Hong Liu.
\newblock Multi-scale spatial temporal graph convolutional network for
  skeleton-based action recognition.
\newblock In {\em Proceedings of the AAAI Conference on Artificial
  Intelligence}, volume~35, pages 1113--1122, 2021.

\bibitem{cheng2020decoupling}
Ke Cheng, Yifan Zhang, Congqi Cao, Lei Shi, Jian Cheng, and Hanqing Lu.
\newblock Decoupling gcn with dropgraph module for skeleton-based action
  recognition.
\newblock In {\em Proceedings of the European Conference on Computer Vision
  (ECCV)}, 2020.

\bibitem{cheng2020skeleton}
Ke Cheng, Yifan Zhang, Xiangyu He, Weihan Chen, Jian Cheng, and Hanqing Lu.
\newblock Skeleton-based action recognition with shift graph convolutional
  network.
\newblock In {\em Proceedings of the IEEE/CVF Conference on Computer Vision and
  Pattern Recognition}, pages 183--192, 2020.

\bibitem{chi2022infogcn}
Hyung-gun Chi, Myoung~Hoon Ha, Seunggeun Chi, Sang~Wan Lee, Qixing Huang, and
  Karthik Ramani.
\newblock Infogcn: Representation learning for human skeleton-based action
  recognition.
\newblock In {\em Proceedings of the IEEE/CVF Conference on Computer Vision and
  Pattern Recognition}, pages 20186--20196, 2022.

\bibitem{duvenaud2015convolutional}
David~K Duvenaud, Dougal Maclaurin, Jorge Iparraguirre, Rafael Bombarell,
  Timothy Hirzel, Al{\'a}n Aspuru-Guzik, and Ryan~P Adams.
\newblock Convolutional networks on graphs for learning molecular fingerprints.
\newblock {\em Advances in neural information processing systems}, 28, 2015.

\bibitem{hamilton2017inductive}
Will Hamilton, Zhitao Ying, and Jure Leskovec.
\newblock Inductive representation learning on large graphs.
\newblock {\em Advances in neural information processing systems}, 30, 2017.

\bibitem{he2016deep}
Kaiming He, Xiangyu Zhang, Shaoqing Ren, and Jian Sun.
\newblock Deep residual learning for image recognition.
\newblock In {\em Proceedings of the IEEE Conference on Computer Vision and
  Pattern Recognition}, pages 770--778, 2016.

\bibitem{hu2021learning}
Qingyong Hu, Bo Yang, Linhai Xie, Stefano Rosa, Yulan Guo, Zhihua Wang, Niki
  Trigoni, and Andrew Markham.
\newblock Learning semantic segmentation of large-scale point clouds with
  random sampling.
\newblock {\em IEEE Transactions on Pattern Analysis and Machine Intelligence},
  2021.

\bibitem{jang2016categorical}
Eric Jang, Shixiang Gu, and Ben Poole.
\newblock Categorical reparameterization with gumbel-softmax.
\newblock {\em arXiv preprint arXiv:1611.01144}, 2016.

\bibitem{kay2017kinetics}
Will Kay, Joao Carreira, Karen Simonyan, Brian Zhang, Chloe Hillier, Sudheendra
  Vijayanarasimhan, Fabio Viola, Tim Green, Trevor Back, Paul Natsev, et~al.
\newblock The kinetics human action video dataset.
\newblock {\em arXiv preprint arXiv:1705.06950}, 2017.

\bibitem{ke2022towards}
Lipeng Ke, Kuan-Chuan Peng, and Siwei Lyu.
\newblock Towards to-at spatio-temporal focus for skeleton-based action
  recognition.
\newblock {\em arXiv preprint arXiv:2202.02314}, 2022.

\bibitem{kipf2018neural}
Thomas Kipf, Ethan Fetaya, Kuan-Chieh Wang, Max Welling, and Richard Zemel.
\newblock Neural relational inference for interacting systems.
\newblock In {\em International Conference on Machine Learning}, pages
  2688--2697. PMLR, 2018.

\bibitem{korban2020ddgcn}
Matthew Korban and Xin Li.
\newblock Ddgcn: A dynamic directed graph convolutional network for action
  recognition.
\newblock In {\em European Conference on Computer Vision}, pages 761--776.
  Springer, 2020.

\bibitem{lee2022hierarchically}
Jungho Lee, Minhyeok Lee, Dogyoon Lee, and Sangyoon Lee.
\newblock Hierarchically decomposed graph convolutional networks for
  skeleton-based action recognition.
\newblock {\em arXiv preprint arXiv:2208.10741}, 2022.

\bibitem{li2019deepgcns}
Guohao Li, Matthias Müller, Ali Thabet, and Bernard Ghanem.
\newblock Deepgcns: Can gcns go as deep as cnns?
\newblock In {\em The IEEE International Conference on Computer Vision (ICCV)},
  2019.

\bibitem{liu2019ntu}
Jun Liu, Amir Shahroudy, Mauricio Perez, Gang Wang, Ling-Yu Duan, and Alex~C
  Kot.
\newblock Ntu rgb+ d 120: A large-scale benchmark for 3d human activity
  understanding.
\newblock {\em IEEE Transactions on Pattern Analysis and Machine Intelligence},
  42(10):2684--2701, 2019.

\bibitem{liu2017skeleton}
Jun Liu, Amir Shahroudy, Dong Xu, Alex~C Kot, and Gang Wang.
\newblock Skeleton-based action recognition using spatio-temporal lstm network
  with trust gates.
\newblock {\em IEEE transactions on pattern analysis and machine intelligence},
  40(12):3007--3021, 2017.

\bibitem{liu2017global}
Jun Liu, Gang Wang, Ping Hu, Ling-Yu Duan, and Alex~C Kot.
\newblock Global context-aware attention lstm networks for 3d action
  recognition.
\newblock In {\em Proceedings of the IEEE conference on computer vision and
  pattern recognition}, pages 1647--1656, 2017.

\bibitem{liu2020disentangling}
Ziyu Liu, Hongwen Zhang, Zhenghao Chen, Zhiyong Wang, and Wanli Ouyang.
\newblock Disentangling and unifying graph convolutions for skeleton-based
  action recognition.
\newblock In {\em Proceedings of the IEEE/CVF Conference on Computer Vision and
  Pattern Recognition}, pages 143--152, 2020.

\bibitem{loshchilov2016sgdr}
Ilya Loshchilov and Frank Hutter.
\newblock Sgdr: Stochastic gradient descent with warm restarts.
\newblock {\em arXiv preprint arXiv:1608.03983}, 2016.

\bibitem{monti2017geometric}
Federico Monti, Davide Boscaini, Jonathan Masci, Emanuele Rodola, Jan Svoboda,
  and Michael~M Bronstein.
\newblock Geometric deep learning on graphs and manifolds using mixture model
  cnns.
\newblock In {\em Proceedings of the IEEE conference on computer vision and
  pattern recognition}, pages 5115--5124, 2017.

\bibitem{niepert2016learning}
Mathias Niepert, Mohamed Ahmed, and Konstantin Kutzkov.
\newblock Learning convolutional neural networks for graphs.
\newblock In {\em International conference on machine learning}, pages
  2014--2023. PMLR, 2016.

\bibitem{shahroudy2016ntu}
Amir Shahroudy, Jun Liu, Tian-Tsong Ng, and Gang Wang.
\newblock Ntu rgb+ d: A large scale dataset for 3d human activity analysis.
\newblock In {\em Proceedings of the IEEE Conference on Computer Vision and
  Pattern Recognition}, pages 1010--1019, 2016.

\bibitem{shi2019skeleton}
Lei Shi, Yifan Zhang, Jian Cheng, and Hanqing Lu.
\newblock Skeleton-based action recognition with directed graph neural
  networks.
\newblock In {\em Proceedings of the IEEE/CVF Conference on Computer Vision and
  Pattern Recognition}, pages 7912--7921, 2019.

\bibitem{shi2019two}
Lei Shi, Yifan Zhang, Jian Cheng, and Hanqing Lu.
\newblock Two-stream adaptive graph convolutional networks for skeleton-based
  action recognition.
\newblock In {\em Proceedings of the IEEE/CVF Conference on Computer Vision and
  Pattern Recognition}, pages 12026--12035, 2019.

\bibitem{shi2020skeleton}
Lei Shi, Yifan Zhang, Jian Cheng, and Hanqing Lu.
\newblock Skeleton-based action recognition with multi-stream adaptive graph
  convolutional networks.
\newblock {\em IEEE Transactions on Image Processing}, 29:9532--9545, 2020.

\bibitem{si2019attention}
Chenyang Si, Wentao Chen, Wei Wang, Liang Wang, and Tieniu Tan.
\newblock An attention enhanced graph convolutional lstm network for
  skeleton-based action recognition.
\newblock In {\em Proceedings of the IEEE/CVF Conference on Computer Vision and
  Pattern Recognition}, pages 1227--1236, 2019.

\bibitem{song2022constructing}
Yi-Fan Song, Zhang Zhang, Caifeng Shan, and Liang Wang.
\newblock Constructing stronger and faster baselines for skeleton-based action
  recognition.
\newblock {\em IEEE Transactions on Pattern Analysis and Machine Intelligence},
  2022.

\bibitem{veeriah2015differential}
Vivek Veeriah, Naifan Zhuang, and Guo-Jun Qi.
\newblock Differential recurrent neural networks for action recognition.
\newblock In {\em Proceedings of the IEEE international conference on computer
  vision}, pages 4041--4049, 2015.

\bibitem{wang2014cross}
Jiang Wang, Xiaohan Nie, Yin Xia, Ying Wu, and Song-Chun Zhu.
\newblock Cross-view action modeling, learning and recognition.
\newblock In {\em Proceedings of the IEEE Conference on Computer Vision and
  Pattern Recognition}, pages 2649--2656, 2014.

\bibitem{wang2016temporal}
Limin Wang, Yuanjun Xiong, Zhe Wang, Yu Qiao, Dahua Lin, Xiaoou Tang, and
  Luc~Van Gool.
\newblock Temporal segment networks: Towards good practices for deep action
  recognition.
\newblock In {\em European conference on computer vision}, pages 20--36.
  Springer, 2016.

\bibitem{wang2018non}
Xiaolong Wang, Ross Girshick, Abhinav Gupta, and Kaiming He.
\newblock Non-local neural networks.
\newblock In {\em Proceedings of the IEEE conference on computer vision and
  pattern recognition}, pages 7794--7803, 2018.

\bibitem{Xiang_2021_ICCV}
Tiange Xiang, Chaoyi Zhang, Yang Song, Jianhui Yu, and Weidong Cai.
\newblock Walk in the cloud: Learning curves for point clouds shape analysis.
\newblock In {\em Proceedings of the IEEE/CVF International Conference on
  Computer Vision (ICCV)}, pages 915--924, October 2021.

\bibitem{yan2018spatial}
Sijie Yan, Yuanjun Xiong, and Dahua Lin.
\newblock Spatial temporal graph convolutional networks for skeleton-based
  action recognition.
\newblock In {\em Thirty-second AAAI Conference on Artificial Intelligence},
  2018.

\bibitem{yang2019modeling}
Jiancheng Yang, Qiang Zhang, Bingbing Ni, Linguo Li, Jinxian Liu, Mengdie Zhou,
  and Qi Tian.
\newblock Modeling point clouds with self-attention and gumbel subset sampling.
\newblock In {\em Proceedings of the IEEE/CVF conference on computer vision and
  pattern recognition}, pages 3323--3332, 2019.

\bibitem{ying2018hierarchical}
Zhitao Ying, Jiaxuan You, Christopher Morris, Xiang Ren, Will Hamilton, and
  Jure Leskovec.
\newblock Hierarchical graph representation learning with differentiable
  pooling.
\newblock {\em Advances in neural information processing systems}, 31, 2018.

\bibitem{zhang2020semantics}
Pengfei Zhang, Cuiling Lan, Wenjun Zeng, Junliang Xing, Jianru Xue, and Nanning
  Zheng.
\newblock Semantics-guided neural networks for efficient skeleton-based human
  action recognition.
\newblock In {\em Proceedings of the IEEE/CVF Conference on Computer Vision and
  Pattern Recognition}, pages 1112--1121, 2020.

\end{thebibliography}
}

\end{document}